\documentclass[11pt]{article}
\usepackage[preprint]{acl}
\usepackage{times}
\usepackage{latexsym}
\usepackage[T1]{fontenc}
\usepackage[utf8]{inputenc}
\usepackage{microtype}
\IfFileExists{inconsolata.sty}
  {\usepackage{inconsolata}}
  {\usepackage{courier}}
\usepackage{graphicx}
\usepackage{booktabs}
\usepackage{amsmath}
\usepackage{enumitem}
\usepackage{caption}
\usepackage{float}
\usepackage{placeins}

\title{Do LLMs Reproduce the Age Gap in Loneliness?\\
Evaluating Survey Simulation against European Survey Data}

\author{
  Yuliya Li$^{1}$ \quad
  Achim Rettinger$^{1}$ \quad
  Dominik Beyer$^{2}$\\
  $^{1}$Trier University, Germany\\
  $^{2}$FZI Research Center for Information Technology,
  Karlsruhe, Germany\\
  \texttt{\{s2yyliii,rettinger\}@uni-trier.de}\\
  \texttt{beyer@fzi.de}
}

\begin{document}
\maketitle
\pagestyle{plain}
\thispagestyle{plain}

\begin{abstract}
Large language models (LLMs) are often used to simulate survey
respondents, but matching survey averages does not establish
accurate demographic differences. We test whether LLM outputs
reproduce within-country age gaps in loneliness using the 2022 EU
Loneliness Survey, comparing adults aged 18--34 and 65 or older in
France, Italy, Poland, and Sweden. We evaluate simulation with and
without age information and direct distribution estimation against
a probability-based sample, with a non-probability sensitivity
reference. Both human samples show higher mean loneliness among
younger adults. Under age-specified simulation with standard
response-option order, Mistral Small 4 and GPT-5.6 Terra reverse
this ordering under all three prompt framings in three and two
countries, respectively, mainly by overestimating older-adult
means. DeepSeek V4 Flash preserves the ordering everywhere but
underestimates both group means, with errors largely cancelling in
France and Italy. Against the probability-based reference, direct
estimation has lower absolute age-gap error than simulation for
Mistral and Terra in every country. For DeepSeek, simulation has
lower gap error but direct estimation has lower group-mean error;
these comparisons depend on low-validity older-adult simulation
cells. Evaluating LLM outputs as proxies for age groups within
countries therefore requires checking both group means and their
contrast against human survey data.
\end{abstract}

\section{Introduction}
\label{sec:introduction}

Large language models (LLMs) are increasingly used as proxies for survey respondents: models receive demographic profiles and answer
as members of the specified groups \citep{argyle2023out}. A profile
that names only a country gives the model no information to distinguish groups within that country. Whether adding a demographic
attribute reproduces observed differences between groups is an empirical question. We test this for age differences in loneliness within four European countries.

Loneliness provides a useful test case. In a representative German sample, \citet{luhmann2016age} found elevated loneliness among young adults as well as the oldest old, a pattern they describe as contrary to stereotypes associating loneliness primarily with old age. In an international online sample, loneliness decreased with age \citep{barreto2021loneliness}. In our four-country survey reference, adults aged 18--34 have higher mean loneliness scores than adults aged 65 or older (Section~4.1). A model that assigns a higher mean to older adults would therefore reverse this observed ordering.

\citet{plum2025identity} include sensitivity to intra-cultural variation
among their criteria for cultural reasoning. This means acknowledging
differences within a cultural context, for example by region or age.
Survey data make a related question measurable: when a model answers
as a member of an age group, do its group means and their difference
match survey estimates from the same country? We address this
question for loneliness, measuring agreement with survey data rather
than cultural reasoning itself.

Models can match overall survey averages while misrepresenting differences between demographic groups \citep{bisbee2024synthetic}. Agreement with
published survey marginals can also be obtained through direct population
estimation, without simulating individual respondents
\citep{doudkin2026marginal}. Our evaluation therefore compares respondent simulation with direct estimation. The older-minus-younger gap is our primary target.
We also report both group means: an equal error in the two means cancels in their difference, so an accurate gap can conceal inaccurate levels.

Related benchmarks evaluate opinions on public issues
\citep{santurkar2023whose,meister2025benchmarking} and interest in
books \citep{meister2025benchmarking}; we instead measure
self-reported loneliness with the three-item UCLA Loneliness Scale
\citep{hughes2004short}. Three elicitation methods are compared against two human samples with different sampling designs.
Evaluating the same outputs against both samples shows whether the size
of model errors and the ranking of elicitation methods depend on which
human sample serves as the reference.

The human data come from the 2022 EU Loneliness Survey. The analysis compares
adults aged 18--34 and 65 or older in France, Italy, Poland, and Sweden. Our main
research question is:

\begin{description}[leftmargin=0pt,itemsep=2pt,topsep=3pt]
  \item[RQ] How closely does age-specified LLM survey simulation
reproduce the direction and magnitude of the observed
older-minus-younger loneliness gap in each country?
\end{description}

\noindent Each country is evaluated separately so that pooled results do not
conceal country-specific errors. Three supporting questions address the
elicitation method and reference data:

\begin{description}[leftmargin=0pt,itemsep=2pt,topsep=3pt]
  \item[SQ1] How does stating the age group in the prompt change the
  mean absolute error of simulated group means, compared with an
  otherwise identical prompt without an age group?

  \item[SQ2] How does absolute age-gap error differ between direct
distribution estimation and respondent simulation, and does
their ranking agree with the comparison of group-mean errors?

 \item[SQ3] How do age-specified simulation gaps vary across
  prompt framings, and how does the human reference change
  the absolute gap errors and ranking of simulation and
  direct estimation?
\end{description}

This paper provides an empirical evaluation of age-group representation in loneliness within four European countries, comparing simulation with direct estimation and assessing agreement in both group means and their contrast. 

Age-specified simulation reproduces the observed age ordering in all four countries only for DeepSeek. Mistral and Terra reverse the ordering under all three prompt framings in three and two countries, respectively, mainly because they overestimate older-adult means. DeepSeek underestimates both group means, with errors largely cancelling in France and Italy. Stating age does not consistently improve group-mean accuracy. Against the probability-based sample, direct estimation has lower absolute age-gap error than simulation for Mistral and Terra in every country. For DeepSeek, simulation has lower gap error but direct estimation has lower group-mean error; these comparisons depend on low-validity older-adult simulation cells. Using the non-probability reference changes the ranking of
simulation and direct estimation by absolute age-gap error in
two of twelve model--country comparisons, both in France. These findings support evaluating demographic differences within countries through both group means and their contrast, alongside output validity.

\section{Related Work}
\label{sec:related-work}

\paragraph{LLMs as survey simulators.}
\citet{argyle2023out} prompt a language model with sociodemographic
backstories of real survey participants. They report that the resulting
``silicon samples'' reproduce subgroup response distributions and
relationships among demographics, attitudes, and reported behavior in
their studies. \citet{santurkar2023whose} find substantial differences
between model opinions and those of US demographic groups. Adding
descriptions of the target groups to the prompts (``steering'') yields
only modest improvements and leaves disparities in alignment across
the tested groups. Adults aged 65 or older are among the groups least
well represented in default outputs, but age groups were not included
in their steering experiment.
\citet{bisbee2024synthetic} find that synthetic responses can approximate
survey averages while varying less than human responses. Their regression
analyses also show that associations between respondent characteristics
and responses often differ from those in the survey, sometimes in sign.
Accurate averages therefore do not guarantee accurate relationships with
demographic attributes, motivating a direct evaluation of the age contrast.

\paragraph{Elicitation and prompting.}
In their distributional-alignment benchmark,
\citet{meister2025benchmarking} report lower average total
variation distance when models directly state opinion
distributions than when they generate a sequence of
30 sampled answers. \citet{doudkin2026marginal}
argues that agreement with published survey marginals does not distinguish
respondent simulation from direct population estimation. Post hoc exploratory analyses by Doudkin suggest some ordering
signal across selected demographic cells despite biased response
levels, mainly under probability elicitation. The author treats
this as a hypothesis rather than evidence of reliable recovery
of subgroup structure. We examine age ordering in a different
domain: self-reported loneliness. Model responses also depend on presentation: choice ordering
and labeling affect survey answers \citep{dominguez2024questioning},
while role-adoption formats and demographic priming affect persona-based
simulations \citep{lutz2025prompt}. Together, these findings motivate
comparing simulation with direct estimation, evaluating both group means
and their difference, and checking sensitivity to prompt framing.

\paragraph{Validity of simulated respondents.}

\citet{munker2026dont} argue that social simulations should be evaluated
by how closely simulated behavior matches observed human behavior in
the setting where the simulation components were fitted.

\citet{schwager2026towards} note that explicit conditioning,
in which a person's attributes are described in the prompt,
is rarely tested against ground truth. In their social-media
prediction experiments, generated biographies become redundant
in the fine-tuned setting when behavioral histories are available.
For psychological questionnaires, \citet{munker2025fingerprinting}
finds that model responses to the Humor Styles Questionnaire do not
recover the four-factor structure observed in the human baseline;
those prompts contained no demographic profiles. We apply this validation logic to survey simulation, comparing
age-specified responses with probability-based survey estimates
in each country.

\section{Methods}
\label{sec:methods}
\subsection{Human Reference Data}
\label{sec:human-reference}

We use data from the 2022 EU Loneliness Survey
\citep{jrc2026euls}, fielded in November and December 2022
\citep{berlingieri2024eu}. The EU4 sample ($n=4{,}029$) was
recruited through probability-based panels in France, Italy, Poland,
and Sweden. The survey designers selected these countries to represent
four EU macro-regions. Panel recruitment used random-digit dialling in
France, Italy, and Poland and address-based sampling in Sweden. Most
respondents answered online; digitally excluded respondents were
interviewed by telephone. We use EU4 as the primary reference because
of its probability-based recruitment and inclusion of digitally excluded
respondents. The survey documentation notes that this design can improve
the representation of older and less-educated people
\citep{berlingieri2024eu}. The EU27 sample ($n=25{,}646$) was recruited
from non-probability online consumer panels across all EU Member States.
Respondents from the same four countries ($n=4{,}008$) provide the
sensitivity reference.

Our outcome is the three-item UCLA Loneliness Scale
\citep{hughes2004short}, using the wording and response coding of the
EU Loneliness Survey questionnaire \citep{jrc2026euls}. The items ask
how often respondents feel that they lack companionship, feel left out,
and feel isolated from others. Responses are coded 1 (``hardly ever or
never''), 2 (``some of the time''), and 3 (``often''). The three items
are summed into a score from 3 to 9, with higher scores indicating greater
loneliness. Mean scores are compared without applying a loneliness cutoff.
The human reference and model conditions target the same UCLA-3 score
definition. Appendix~\ref{app:instrument} lists the variables and codes.

Both samples cover people aged 16 or older. We compare adults aged
18--34 with those aged 65 or older. Ages 16--17 fall outside the adult
population addressed by the prompts; ages 35--64 are excluded by design. The younger band is defined for this study,
motivated by evidence of elevated loneliness in young adulthood
\citep{luhmann2016age}. We define the older band as ages 65--100. Adults aged 65 or older
are also among the groups poorly represented in default model
opinions in \citet{santurkar2023whose}.

Before selecting these groups, we remove records with missing ages or
ages outside 16--100. Within the selected groups, we retain only respondents
with all three UCLA responses in $\{1,2,3\}$ and a finite, positive survey
weight; ``prefer not to say'' (999) counts as item nonresponse and is
excluded. The resulting samples contain 1,720 EU4 and 1,682 EU27
respondents from the four countries.
Appendix~\ref{app:human-audit} reports exclusions at each step.

We use the supplied country-specific weight (\texttt{w\_country}).
The JRC report describes demographic weighting using RIM, with
category merging and weight trimming; EU4 weights additionally
incorporate design weights \citep{berlingieri2024eu}.
The supplied weight is therefore used directly, without multiplying
by an additional design weight. Within each country--age group, the weighted
human mean is
\[
\widehat{\mu}_H = \frac{\sum_i w_i y_i}{\sum_i w_i},
\]
where $y_i$ is respondent $i$'s UCLA-3 score and $w_i$ is their survey
weight. We define the age gap as the older group's weighted mean minus
the younger group's weighted mean. Table~\ref{tab:human-reference}
reports raw and Kish effective sample sizes \citep{kish1965survey},
weighted means, and age gaps with 95\% percentile-bootstrap intervals.
Section~\ref{sec:evaluation} summarizes the uncertainty estimation;
Appendix~\ref{app:evaluation-details} gives the resampling procedure.

\subsection{Models and Conditions}
\label{sec:models-conditions}

We evaluate Mistral Small 4, DeepSeek V4 Flash, and GPT-5.6 Terra under three conditions.

\paragraph{S: age specified.}
The model answers the three UCLA items as a respondent in a specified
country and age group in late 2022. The profile supplies only country,
age group, and survey period.

\paragraph{N: age unspecified.}
N uses the S prompt without the sentence stating the age group, so
the two conditions differ only in age information. Its single country
mean serves as a common prediction for both age groups, without
estimating an age-specific contrast. Comparing S with N tests whether
stating age improves the predicted group means (SQ1).

\paragraph{D: direct estimation.}
The model provides seven probabilities for total UCLA-3 scores 3--9
among respondents who complete all three items in the specified country
and age group in late 2022. These probabilities define the predicted
mean.

\paragraph{Prompt variation.}
We use three prompt framings per condition, varying only the introductory
wording while holding the items, answer labels, numeric codes, and target
population fixed. All prompts are in English to hold language constant
across countries. S and N retain the human questionnaire's ``prefer not
to say'' option (999); these outputs are treated as item nonresponse and
excluded from score estimates. Appendix~\ref{app:proposed-prompts}
provides the matched prompts and collection details.

S asks for one respondent's three item answers; D asks for a population
distribution over total scores. Since both perspective and output format
change, SQ2 compares the complete procedures and cannot attribute their
difference to perspective alone. The comparison follows the distinction
between directly stated distributions and generated answer sequences in
\citet{meister2025benchmarking}. Unlike their 30-answer sequence condition,
S obtains each simulated response in a fresh conversation, and the primary
target is an age-group contrast rather than the full opinion distribution.

We scheduled 50 draws in each of 180 model--country--framing--variant
cells, for 9,000 draws. The five variants are younger and older S,
N, and younger and older D. A draw is one planned model observation;
technical retries remain part of that draw and do not increase the
scheduled count.

\subsection{Evaluation and Uncertainty}
\label{sec:evaluation}

We evaluate each model separately in each country, using EU4 as
the primary reference and matched EU27 as an alternative reference.
Below, $a\in\{y,o\}$ denotes younger and older adults, and
$X\in\{S,D\}$ denotes the elicitation condition.
N provides one country mean without an age index.

\paragraph{Means and age gaps.}
All model estimates are conditional on complete valid outputs.
For S, we average generated UCLA-3 totals; for D, we average
the means implied by accepted probability vectors.
We calculate a mean for each prompt framing and then average
the three framing means with equal weight.
N follows the same procedure without age groups.

Let $\widehat{\mu}_{X,a}$ be the model mean and
$\widehat{\mu}_{H,a}$ the weighted human mean. We define
\begin{align}
\widehat{\Delta}_X
  &= \widehat{\mu}_{X,o}-\widehat{\mu}_{X,y},\\
\widehat{E}_X
  &= \widehat{\Delta}_X-\widehat{\Delta}_H,
\end{align}
where $\widehat{\Delta}_H$ is the human older-minus-younger gap.
Negative gaps indicate higher mean loneliness among younger adults.
A positive gap error means that the model gap is more positive
than the human gap.
We describe ordering as reversed when the model and human gap
point estimates have opposite signs. We separately report whether
this occurs under all three framings.
The primary evaluation compares the S and EU4 gaps,
reporting signed and absolute gap errors.

\paragraph{Group means and condition comparisons.}
Define the signed error for age group $a$ as
$\widehat{e}_{X,a}
=\widehat{\mu}_{X,a}-\widehat{\mu}_{H,a}$.
The signed gap error therefore satisfies
\begin{equation}
\widehat{E}_X
=\widehat{e}_{X,o}-\widehat{e}_{X,y}.
\label{eq:gap-error-components}
\end{equation}
We summarize agreement in score levels using the mean absolute
error across the two age groups:
\begin{equation}
L_X=\frac{1}{2}\sum_{a\in\{y,o\}}
\left|\widehat{e}_{X,a}\right|.
\end{equation}
This summary measures error magnitude but does not identify
which age group contributes more.
For $L_N$, the same country mean is compared with both human
age-group means. Table~\ref{tab:elicitation-comparison} reports
$L_N-L_S$, which is positive when stating age reduces group-mean
error, and $|\widehat{E}_S|-|\widehat{E}_D|$, which is positive
when D has smaller absolute gap error.

\paragraph{Sensitivity and output validity.}
We report S gaps separately for each framing and evaluate
the same combined S and D gaps against both human references.
Before the main run, we set 45 valid outputs as the minimum
for interval reporting in each required framing cell.
A framing cell below this threshold is a \emph{low-validity cell}:
its point estimates are retained, but intervals for quantities
that depend on it are withheld.

\paragraph{Uncertainty.}
We use 10{,}000 bootstrap resamples of human respondents within
sample--country--age groups, retaining each score--weight pair.
These intervals are approximate: they keep the supplied weights
fixed rather than reconstructing the full survey-design variance.
Model-only intervals resample valid outputs within experimental
cells; intervals for model--human discrepancies combine model
and human resampling.
We report pointwise 95\% percentile intervals for human means
and gaps, combined model gaps, and signed EU4 gap errors
\citep{efron1993introduction}.
Absolute errors and differences in absolute error are reported
descriptively, without inferential intervals or hypothesis tests.
Appendix~\ref{app:evaluation-details} gives the estimator,
validity, and resampling details.

\section{Results}
\label{sec:results}
\raggedbottom
\renewcommand{\dblfloatpagefraction}{0.95}
\renewcommand{\floatpagefraction}{0.95}
\makeatletter
\setlength{\@fptop}{0pt}
\setlength{\@fpsep}{12pt}
\setlength{\@fpbot}{0pt plus 1fil}
\setlength{\@dblfptop}{0pt}
\setlength{\@dblfpsep}{12pt}
\setlength{\@dblfpbot}{0pt plus 1fil}
\makeatother

\subsection{Age Gap in Survey Data}
\label{sec:results-human}

Younger adults have higher weighted mean UCLA-3 scores in all four
countries under both human references (Table~\ref{tab:human-reference};
group means in Appendix Figure~\ref{fig:human-means}).
The EU4 older-minus-younger gaps range from $-1.30$ in Sweden to
$-0.89$ in Italy, and all eight human gap intervals lie below zero
(Table~\ref{tab:human-reference}).
The two references agree on direction but differ in magnitude by up
to about $0.5$ points: France's gap is $-1.00$ under EU4 but $-0.52$
under EU27. EU4 remains the primary reference; EU27 tests sensitivity
to this choice.

\begin{table*}[!htbp]
\centering
\small
\setlength{\tabcolsep}{6pt}
\begin{tabular}{@{}lcccccc@{}}
\toprule
 & \multicolumn{2}{c}{$n$ ($n_{\mathrm{eff}}$)}
 & \multicolumn{2}{c}{Weighted mean}
 & \multicolumn{2}{c}{Older-minus-younger gap} \\
\cmidrule(lr){2-3}\cmidrule(lr){4-5}\cmidrule(l){6-7}
Country & 18--34 & 65+ & 18--34 & 65+ & Estimate & 95\% CI \\
\midrule
\multicolumn{7}{@{}l}{\textit{EU4: probability-based sample}} \\
France & 196 (138) & 271 (168) & 5.05 & 4.06
       & $-1.00$ & [$-1.36$, $-0.62$] \\
Italy  & 176 (102) & 263 (163) & 5.09 & 4.19
       & $-0.89$ & [$-1.31$, $-0.49$] \\
Poland & 219 (93)  & 98 (70)   & 5.38 & 4.27
       & $-1.10$ & [$-1.63$, $-0.57$] \\
Sweden & 203 (196) & 294 (290) & 5.25 & 3.95
       & $-1.30$ & [$-1.59$, $-1.02$] \\
\addlinespace
\multicolumn{7}{@{}l}{
  \textit{EU27: non-probability online sample, matched countries}} \\
France & 232 (148) & 202 (133) & 4.91 & 4.39
       & $-0.52$ & [$-0.92$, $-0.12$] \\
Italy  & 198 (122) & 223 (164) & 5.74 & 4.39
       & $-1.35$ & [$-1.79$, $-0.90$] \\
Poland & 262 (193) & 90 (64)   & 5.36 & 4.51
       & $-0.84$ & [$-1.35$, $-0.31$] \\
Sweden & 269 (224) & 206 (163) & 5.74 & 3.95
       & $-1.79$ & [$-2.11$, $-1.46$] \\
\bottomrule
\end{tabular}
\caption{Human reference estimates. $n$: complete respondents with
valid weights; $n_{\mathrm{eff}}$: Kish effective sample size.
Means use the supplied country weights and the UCLA-3 scale (3--9).
Gaps compare ages 65--100 with 18--34; intervals are pointwise
95\% percentile-bootstrap intervals. Displayed estimates are rounded.}
\label{tab:human-reference}
\end{table*}

\subsection{Primary EU4 Model Comparison}
\label{sec:results-eu4}

\begin{table*}[!htbp]
\centering
\small
\setlength{\tabcolsep}{4.5pt}
\begin{tabular}{@{}llcrrrl@{}}
\toprule
Country & Model & Cond. & Mean 18--34 & Mean 65+ & Gap
& Signed gap error [95\% CI] \\
\midrule
France
& Mistral Small 4 & S & 5.09 & 6.02 & 0.93
& 1.92 [1.45, 2.38] \\
& Mistral Small 4 & D & 4.77 & 4.70 & $-0.08$
& \textbf{0.92} [0.52, 1.31] \\
& DeepSeek V4 Flash & S & 4.34 & 3.40$^{\dagger}$
& $-0.93^{\dagger}$ & \textbf{0.06}$^{\dagger}$ [IW] \\
& DeepSeek V4 Flash & D & 4.48 & 4.11 & $-0.37$
& 0.62 [0.25, 1.00] \\
& GPT-5.6 Terra & S & 5.43 & 5.19 & $-0.23$
& 0.76 [0.37, 1.15] \\
& GPT-5.6 Terra & D & 4.95 & 3.99 & $-0.95$
& \textbf{0.04} [$-0.33$, 0.41] \\
\midrule
Italy
& Mistral Small 4 & S & 4.80 & 6.03 & 1.23
& 2.12 [1.65, 2.62] \\
& Mistral Small 4 & D & 4.74 & 4.78 & 0.05
& \textbf{0.94} [0.51, 1.37] \\
& DeepSeek V4 Flash & S & 4.27 & 3.41$^{\dagger}$
& $-0.86^{\dagger}$ & \textbf{0.03}$^{\dagger}$ [IW] \\
& DeepSeek V4 Flash & D & 4.44 & 4.13 & $-0.31$
& 0.58 [0.17, 1.00] \\
& GPT-5.6 Terra & S & 5.33 & 5.98 & 0.65
& 1.54 [1.13, 1.96] \\
& GPT-5.6 Terra & D & 4.95 & 4.10 & $-0.85$
& \textbf{0.04} [$-0.36$, 0.46] \\
\midrule
Poland
& Mistral Small 4 & S & 4.90 & 6.27 & 1.37
& 2.47 [1.87, 3.07] \\
& Mistral Small 4 & D & 4.86 & 4.72 & $-0.14$
& \textbf{0.96} [0.42, 1.50] \\
& DeepSeek V4 Flash & S & 4.37 & 3.74$^{\dagger}$
& $-0.63^{\dagger}$ & \textbf{0.48}$^{\dagger}$ [IW] \\
& DeepSeek V4 Flash & D & 4.58 & 4.18 & $-0.41$
& 0.70 [0.17, 1.24] \\
& GPT-5.6 Terra & S & 5.61 & 6.00 & 0.39
& 1.49 [0.95, 2.03] \\
& GPT-5.6 Terra & D & 4.77 & 4.18 & $-0.59$
& \textbf{0.51} [$-0.02$, 1.05] \\
\midrule
Sweden
& Mistral Small 4 & S & 4.76 & 4.80 & 0.03
& 1.33 [0.93, 1.75] \\
& Mistral Small 4 & D & 4.68 & 4.63 & $-0.06$
& \textbf{1.25} [0.93, 1.55] \\
& DeepSeek V4 Flash & S & 4.11 & 3.37$^{\dagger}$
& $-0.74^{\dagger}$ & \textbf{0.56}$^{\dagger}$ [IW] \\
& DeepSeek V4 Flash & D & 4.21 & 3.98 & $-0.23$
& 1.07 [0.78, 1.37] \\
& GPT-5.6 Terra & S & 5.04 & 4.57 & $-0.47$
& 0.83 [0.51, 1.15] \\
& GPT-5.6 Terra & D & 4.54 & 3.85 & $-0.69$
& \textbf{0.61} [0.33, 0.90] \\
\bottomrule
\end{tabular}

\caption{EU4 estimates under S (age-specified simulation) and D
(direct estimation), averaged equally across the three framings.
Gap: older minus younger; gap error: model gap minus EU4 human gap.
Intervals are pointwise 95\% bootstrap intervals combining model
and fixed-weight human resampling, conditional on the three selected framings. Bold marks the smaller absolute
gap error in each model--country S/D pair.
$^{\dagger}$Depends on a low-validity cell; IW: interval withheld.}
\label{tab:eu4-model-results}
\end{table*}

\begin{figure}[t]
\centering
\includegraphics[width=\columnwidth]{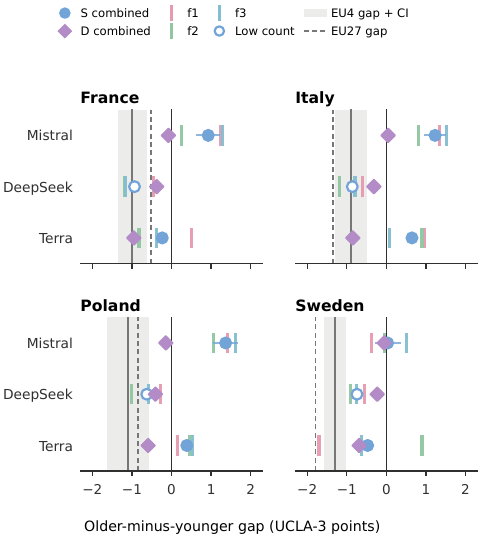}
\caption{Older-minus-younger UCLA-3 gaps; negative values indicate
higher mean loneliness among younger adults. S and D estimates
average the three framings equally. Bars show model-only pointwise 95\% intervals conditional on
the three selected framings. Hollow circles depend on a low-validity cell
($<45$ valid outputs); dependent intervals are withheld.}
\label{fig:model-gaps}
\end{figure}
Age-specified simulation reproduces the observed age ordering in all
four countries for only one of the three models
(RQ; Figure~\ref{fig:model-gaps}). Mistral's combined S gaps are positive in all four countries. In
France, Italy, and Poland they range from $0.93$ to $1.37$, where
every human gap is negative, and are positive under every framing;
Sweden has a near-zero combined gap ($0.03$) with framing-specific
gaps of both signs. Terra reverses the ordering in Italy and Poland and retains the
direction in France and Sweden, but with much smaller gaps than
observed ($-0.23$ and $-0.47$, against $-1.00$ and $-1.30$). Only
DeepSeek preserves the direction everywhere, with signed gap errors
of $0.03$--$0.56$.

\begin{figure}[!t]
\centering
\includegraphics[width=\columnwidth]{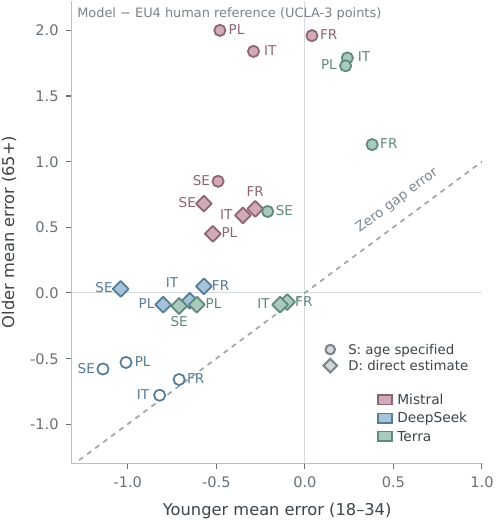}

\caption{Point estimates of errors in the younger and older group
means against EU4, one point per model, country, and method. A point's
height above the dashed diagonal equals its signed gap error
(Equation~\eqref{eq:gap-error-components}); only points near the
origin match both group means. Hollow markers depend on a low-validity
cell. FR, IT, PL, SE: France, Italy, Poland, Sweden.}
\label{fig:error-decomposition}
\end{figure}

The group means locate the discrepancies
(Table~\ref{tab:eu4-model-results};
Figure~\ref{fig:error-decomposition}).
Under S, Mistral and Terra overestimate the older mean in every
country, by approximately 0.6--2.0 points, while their younger
means differ from the human estimates by about 0.5 points or less.
Older-mean overestimation therefore accounts for most of their
positive gap errors
(Equation~\eqref{eq:gap-error-components}).
DeepSeek instead underestimates both means in every country.
These errors nearly cancel in the gap in France and Italy,
yielding gap errors of only $0.06$ and $0.03$.

These DeepSeek S estimates also contain the study's only
low-validity cells. Of 9,000 scheduled draws, 8,911 were valid, but
four DeepSeek older-adult S cells, all under framing~2, had only
41--44 valid outputs. Their point estimates are retained, but
dependent intervals, including all four combined DeepSeek S
gap-error intervals, are withheld (Appendix~\ref{app:model-validity}).

\subsection{Effect of Elicitation Method}
\label{sec:results-elicitation}

Against EU4, D has lower absolute gap error than S in every country
for Mistral and Terra (SQ2; Table~\ref{tab:elicitation-comparison}).
Its displayed group-mean error is also lower, except for a tie for
Terra in Sweden. The two criteria rank the procedures in opposite
order only for DeepSeek, whose S estimates depend on low-validity
cells: S has smaller gap error in every country, while D has
smaller group-mean error. In France, for example, S misses the
gap by $0.06$ but the group means by $0.69$ on average, whereas
D misses the gap by $0.62$ but the group means by $0.31$.
Smaller group-mean errors therefore do not guarantee a smaller gap
error.

\begin{table}[!t]
\centering
\small
\setlength{\tabcolsep}{3pt}
\begin{tabular*}{\columnwidth}{@{\extracolsep{\fill}}lrrrrr@{}}
\toprule
Model & $L_N$ & $L_S$ & $L_D$
& \shortstack{$L_N$\\$-L_S$}
& \shortstack{$|E_S|$\\$-|E_D|$} \\
\midrule
\multicolumn{6}{@{}l}{\textit{France}} \\
Mistral & 0.50 & 1.00 & \textbf{0.46} & $-0.50$ & 1.00 \\
DeepSeek & 0.89 & 0.69$^{\dagger}$ & \textbf{0.31} & 0.20$^{\dagger}$ & $-0.56^{\dagger}$ \\
Terra & 0.54 & 0.75 & \textbf{0.09} & $-0.22$ & 0.72 \\
\addlinespace[2pt]
\multicolumn{6}{@{}l}{\textit{Italy}} \\
Mistral & \textbf{0.45} & 1.06 & 0.47 & $-0.62$ & 1.18 \\
DeepSeek & 0.97 & 0.80$^{\dagger}$ & \textbf{0.35} & 0.17$^{\dagger}$ & $-0.55^{\dagger}$ \\
Terra & 1.01 & 1.02 & \textbf{0.12} & 0.00 & 1.50 \\
\addlinespace[2pt]
\multicolumn{6}{@{}l}{\textit{Poland}} \\
Mistral & 0.55 & 1.24 & \textbf{0.48} & $-0.68$ & 1.51 \\
DeepSeek & 1.22 & 0.77$^{\dagger}$ & \textbf{0.44} & 0.45$^{\dagger}$ & $-0.22^{\dagger}$ \\
Terra & 0.55 & 0.98 & \textbf{0.35} & $-0.43$ & 0.98 \\
\addlinespace[2pt]
\multicolumn{6}{@{}l}{\textit{Sweden}} \\
Mistral & 0.65 & 0.67 & \textbf{0.62} & $-0.02$ & 0.09 \\
DeepSeek & 1.05 & 0.87$^{\dagger}$ & \textbf{0.54} & 0.18$^{\dagger}$ & $-0.51^{\dagger}$ \\
Terra & 0.65 & \textbf{0.41} & \textbf{0.41} & 0.24 & 0.21 \\
\bottomrule
\end{tabular*}
\caption{EU4 error comparison. $L$: mean absolute error across
the two age-group means; $E$: signed gap error.
Positive $L_N-L_S$ favors S over N; positive
$|E_S|-|E_D|$ favors D over S.
Bold marks the lowest displayed $L$ per row, including rounding ties.
$^{\dagger}$Depends on a low-validity cell.}
\label{tab:elicitation-comparison}
\end{table}

Equation~\eqref{eq:gap-error-components} locates the D errors. For
DeepSeek and Terra, the older-adult D means lie within about $0.1$
points of the human estimates in all four countries, so their gap
errors arise mainly from underestimated younger means; under S, overestimation of older-adult means contributes most to the gap errors for Mistral and Terra
(Section~\ref{sec:results-eu4}). Terra D approximates both means and
the gap in France and Italy (gap errors of $0.04$; group-mean errors
of $0.09$ and $0.12$), but its younger means fall further below the
human estimates in Poland and Sweden, where its gap errors rise to
$0.51$ and $0.61$. Mistral D overestimates the older mean and
underestimates the younger mean in every country, so its two errors
add in the gap rather than cancel.

Specifying age reduces group-mean error relative to N (SQ1) for
DeepSeek in all four countries and increases it for Mistral in all
four; for Terra, it reduces error in Sweden, increases it in France
and Poland, and makes little difference in Italy. Because N gives one
prediction for both age groups, its group-mean error cannot fall below
half the human gap, $|\widehat{\Delta}_H|/2$; any prediction between
the two human means reaches this floor (Appendix~\ref{app:n-baseline}).

Age-specific
predictions are not bound by this floor, yet S falls below it in only
one of twelve model--country cells (Terra in Sweden), whereas D does
so in eleven.

\subsection{Framing and Reference Sensitivity}
\label{sec:results-sensitivity}

For Mistral in France, Italy, and Poland, and Terra in Italy and
Poland, S gap point estimates are positive under all three framings
(SQ3; Table~\ref{tab:framing-gaps}).
Mistral's near-zero combined Sweden gap instead averages
framing-specific estimates with different signs. Where Terra retains
the observed direction, its framings disagree in sign: its gaps range
from $-0.82$ to $0.50$ in France and from $-1.70$ to $0.90$ in Sweden.
DeepSeek's gaps are negative in all 12 country--framing combinations,
but its close France estimate partly reflects cancellation across
framings: signed errors of $0.545$, $-0.195$, and $-0.165$ average to
$0.062$.

In a separate S-only screen in France (framing f1; six draws per
cell), reversing response-option order left Mistral's gap point estimate positive and DeepSeek's negative, but changed Terra's
from positive to negative (Appendix~\ref{app:order-results}).

Using EU27 preserves the S/D gap-error ranking in 10 of 12
model--country comparisons. Both changes occur in France, whose
EU27 gap ($-0.52$) is about half its EU4 gap ($-1.00$). Against EU27, DeepSeek D ($-0.37$) and Terra S ($-0.23$)
have smaller absolute gap errors than the alternative procedure,
reversing their EU4 rankings
(Table~\ref{tab:reference-sensitivity}).

\section{Discussion and Conclusion}
\label{sec:discussion}

Providing age information in a simulation prompt does not reliably
recover the observed age ordering. Mistral and Terra reverse it
under all three framings in three and two countries, respectively,
mainly because they overestimate older-adult means while their
younger-adult means stay within about 0.5 points of the survey
estimates. This pattern is consistent with the stereotype, described
by \citet{luhmann2016age}, that associates loneliness primarily with
old age. However, our design cannot determine whether the errors
arise from such a stereotype or from another mechanism. Varying the
age wording while holding the model deployment fixed would test
whether older-mean overestimation persists across alternative age
descriptions.

Under S, DeepSeek alone preserves the observed direction in every
country and has the smallest absolute EU4 gap error in each country,
yet it underestimates both group means. A small gap error does not
imply small errors in either group mean
(Equation~\eqref{eq:gap-error-components}): in France and Italy,
DeepSeek's similar errors in the two means partly cancel in their
difference. Its close France estimate also partly reflects
cancellation across framings, and all four older-adult S estimates
depend on low-validity cells.

Against EU4, direct estimation has lower absolute age-gap error
than simulation for Mistral and Terra in every country. Its
displayed group-mean error is also lower, except for a tie for
Terra in Sweden. For DeepSeek, simulation has lower gap error
but direct estimation has lower group-mean error. These DeepSeek
comparisons depend on low-validity simulation cells.

For Mistral and Terra, the advantage of direct estimation is
qualitatively consistent with \citet{meister2025benchmarking},
who find lower average distributional error from directly stated
distributions than from sequences of sampled answers. Our study
instead samples each simulated response separately and evaluates
group means and age gaps, so the evaluation targets and sampling
procedures differ.

An accurate age gap is insufficient evidence that LLM outputs
represent age groups within a country: errors in the group means
can cancel. Agreement on these aggregate measures also does not
establish valid simulation of individual respondents
\citep{doudkin2026marginal}. Evaluations should therefore report
both group means and their contrast against human survey data,
alongside framing-specific estimates, the reference sample,
and output validity.

\section*{Limitations}
\label{sec:limitations}

\paragraph{Selection of valid outputs.}
All estimates are conditional on valid outputs. DeepSeek's
older-adult framing-2 cells had lower validity (85\% vs.\ 99\%
overall), and several of the excluded malformed outputs contained
low-score answers (Appendix~\ref{app:model-validity}). Exclusion
may therefore depend on response content. Withholding intervals
for these cells does not remove this possible selection bias.

\paragraph{Human reference and uncertainty.}

Residual underrepresentation after weighting, notably of the oldest
age category in Poland's EU27 sample \citep{berlingieri2024eu}, may
affect the reference means and gaps, as may complete-case filtering. Our fixed-weight bootstrap only
approximates uncertainty: it does not reconstruct the full
survey-design variance or account for all coverage, nonresponse,
and measurement error. EU27 serves as a non-probability
sensitivity reference; its differences from EU4 cannot be
attributed to sampling design alone, since recruitment and
administration also differ.

\paragraph{Scope and interpretation.}
We study observed UCLA-3 scores in four countries in 2022, not
individual psychological equivalence. The two age bands omit
middle adulthood and combine ages 65--100, and their
cross-sectional contrast does not identify ageing effects. Our
English prompts and brief country--age profiles differ from the
multilingual human survey, and we did not test measurement
invariance.

Our findings are limited to the tested deployments, decoding
settings, and framings. The S--N comparison does not separate age
information from its wording or test other demographic
attributes; the S--D comparison changes both perspective and
output format. The small answer-order screen cannot establish
presentation invariance (Appendix~\ref{app:order-results}).
Finally, the survey predates our data collection and
training-data exposure is unknown, so agreement cannot be
interpreted as prediction of unseen survey outcomes.

\section*{Ethical Considerations}

This study analyzes released, anonymized survey data from the
European Commission's Joint Research Centre
\citep{jrc2026euls}; no new responses were collected from human
participants. Generated responses are model outputs rather than
observations of people's experiences. Treating unvalidated
simulations as population evidence could misrepresent age groups
and their needs; their use requires validation for the intended
population and purpose.

\bibliography{custom}

\clearpage
\appendix
\raggedbottom
 
\section{Human Reference Details}
\label{app:figures}
 
\subsection{Filtering and Weights}
\label{app:human-audit}
 
Table~\ref{tab:human-audit} reports the respondents remaining after
each filtering step.
 
\begin{table}[!htbp]
\centering
\small
\begin{tabular}{@{}lrr@{}}
\toprule
Stage & EU4 & EU27 \\
\midrule
Released sample    & 4,029 & 25,646 \\
Four countries     & 4,029 & 4,008 \\
Valid age, 16--100  & 4,029 & 4,006 \\
Target age groups  & 1,748 & 1,727 \\
Complete UCLA-3    & 1,720 & 1,682 \\
Valid weight      & 1,720 & 1,682 \\
\bottomrule
\end{tabular}
\caption{Respondents remaining after each filtering step.
EU27 is restricted to the four EU4 countries before age and
response exclusions.}
\label{tab:human-audit}
\end{table}
 
Kish effective sample size is
$n_{\mathrm{eff}}=(\sum_i w_i)^2/\sum_i w_i^2$.
It describes weight concentration rather than representativeness
or the full survey-design effect.

\subsection{Variables and Coding}
\label{app:instrument}
 
Both samples use \texttt{loneliness\_ucla\_a},
\texttt{loneliness\_ucla\_b}, and \texttt{loneliness\_ucla\_c}
for lack of companionship, feeling left out, and feeling isolated,
respectively. Country codes are France 10, Italy 15, Poland 21, and Sweden 27.
 
\begin{figure}[!htbp]
\centering
\includegraphics[width=\linewidth]{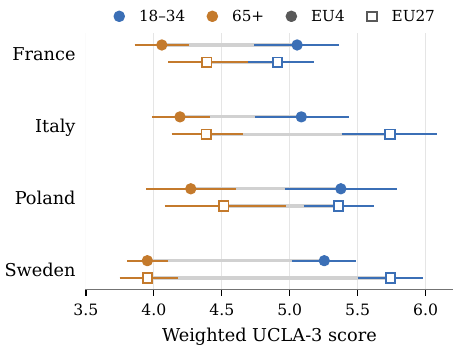}
\caption{Weighted human UCLA-3 means for ages 18--34 and 65--100.
Circles: EU4; squares: matched EU27. Bars show pointwise 95\%
percentile-bootstrap intervals for the means, not the spread of
individual scores.}
\label{fig:human-means}
\end{figure}
 
\section{Prompts and Collection}
\label{app:proposed-prompts}
\subsection{Prompt Construction}
\label{app:prompt-construction}
 
The examples below show framing f1 for France. No human reference
estimates are included in the prompts. Each request uses the system
message: ``Follow the survey instructions and return only the requested
JSON.'' The user message concatenates the introduction, item block,
and output instructions, separated by blank lines.
 
The shared item block below is inserted at \texttt{[ITEMS]} in each
prompt. \texttt{[AGE]} is replaced by ``18 to 34'' or ``65 or older''.
These substitutions specify S-younger, S-older, N, D-younger, and
D-older for this framing. 
 
\paragraph{Shared item block.}
Please indicate how often you feel each of the following:
\begin{enumerate}[leftmargin=*,itemsep=0pt,topsep=3pt]
\item Feel that you lack companionship
\item Feel left out
\item Feel isolated from others
\end{enumerate}
Response codes: 1 = hardly ever or never; 2 = some of the time;
3 = often; 999 = prefer not to say.
The model requests retain the fixed item order shown above.
 
\noindent\begin{minipage}{\linewidth}
\paragraph{S: age specified.}
\begin{quote}\small\raggedright
Answer the following survey as one adult aged 18 or older residing
in France in late 2022. You are aged \texttt{[AGE]}.
 
\texttt{[ITEMS]}
 
Return only a JSON array of three integer response codes in item order.
Use one of 1, 2, 3, or 999 for each item. Do not add explanations.
\end{quote}
\end{minipage}
 
\paragraph{N: age unspecified.}
Use the S prompt with the sentence ``You are aged \texttt{[AGE]}.''
removed. All other text remains identical.
 
\paragraph{D: direct estimation.}
\begin{quote}\small\raggedright
Estimate the distribution of total scores for adults aged
\texttt{[AGE]} residing in France in late 2022 who provide responses
coded 1, 2, or 3 to all three items below.
 
\texttt{[ITEMS]}
 
The total score is the sum of the three item codes and ranges from
3 to 9. Respondents selecting 999 on any item are outside this
complete-response population.
Return only a JSON object with keys ``3'', ``4'', ``5'', ``6'', ``7'',
``8'', and ``9''. Each value must be the estimated proportion with
that total score, between 0 and 1. The seven values must sum to 1.
Do not add explanations.
\end{quote}
 
\paragraph{Framings f2 and f3.}
Only the introduction changes. Country and age placeholders are replaced
by the corresponding cell values. S appends ``You are aged
\texttt{[AGE]}.'' to its introduction; N omits that sentence.
The item block and output instructions remain identical.
 
\begin{quote}\small\raggedright
\textbf{f2, S/N:} Take the perspective of one adult aged 18 or older
residing in \texttt{[COUNTRY]} in late 2022 and answer the following survey.
 
\textbf{f3, S/N:} Complete the following survey from the perspective of
one adult aged 18 or older residing in \texttt{[COUNTRY]} in late 2022.
 
\textbf{f2, D:} Provide your estimate of the distribution of total scores
for adults aged \texttt{[AGE]} residing in \texttt{[COUNTRY]} in late 2022
who provide responses coded 1, 2, or 3 to all three items below.
 
\textbf{f3, D:} Give an estimate of how total scores are distributed
among adults aged \texttt{[AGE]} residing in \texttt{[COUNTRY]} in late 2022
who provide responses coded 1, 2, or 3 to all three items below.
\end{quote}
 
\subsection{Deployment and Collection Controls}
\label{app:deployment}
 
We evaluated Mistral Small 4 and DeepSeek V4 Flash through a
university-hosted endpoint, and GPT-5.6 Terra through the OpenAI API.
The corresponding identifiers were
\texttt{mistral-\allowbreak small-4-119b-a8b},
\texttt{deepseek-\allowbreak v4-flash}, and \texttt{gpt-5.6-terra}.
All three used temperature 0.7 and an output-token cap of 256, with
model reasoning or ``thinking'' modes disabled where the endpoint
exposed them. Dated checkpoint revisions were not exposed by the endpoints.
 
Each draw used a fresh conversation without retrieval or tools, and
JSON was requested in the prompt rather than imposed by constrained
decoding. Experimental cells were interleaved within each model.
Each draw had a distinct seed, which technical retries reused.
 
The main design comprised 180 cells with 50 scheduled draws each.
N contributed one cell per model--country--framing combination
(36 cells, 1{,}800 draws), shared across the two age comparisons. Decoding settings and the
system prompt were fixed across S, N, and D, subject to API
constraints.
 
Before the main run, we tested the prompts in France
(540 scheduled draws) to assess variation in model answers and
collection feasibility. These checks informed the choice of
50 draws per cell, with a target model-generation standard error
of at most 0.15 UCLA-3 points for the combined S gap.
The allocation was not chosen to match the human estimates.
Draws from both the pilot and the answer-order screen
(Appendix~\ref{app:order-results}) are excluded from the main estimates.
 
\section{Evaluation Details}
 
\label{app:evaluation-details}
\subsection{Model Estimators and Validity}

Model and country indices are omitted. Let $t\in\{1,2,3\}$
index framings, $r$ index valid outputs, and $n_{X,a,t}$ count
valid outputs in condition $X$ and age group $a$.
For S, $z_{r,a,t}$ is the total UCLA-3 score, and
$\widehat\mu_{S,a,t}$ is its arithmetic mean over valid outputs
within the framing. For D, $p_{r,a,t}(s)$ is the accepted probability
of total score $s$. We calculate D's mean and the equal-framing average as
\begin{align}
\widehat\mu_{D,a,t}
  &= \frac{1}{n_{D,a,t}}\sum_r
     \sum_{s=3}^{9}s\,p_{r,a,t}(s),\\
\widehat\mu_{X,a}
  &= \frac13\sum_{t=1}^{3}\widehat\mu_{X,a,t}.
\end{align}

N uses the S estimator and equal-framing average without an age
index. Framings retain equal weight when valid counts differ.
A combined estimate is unavailable if a required cell is empty.
Evaluation uses unrounded estimates.

The original analysis plan also included score dispersion as
a secondary diagnostic; this report focuses on group means
and age gaps and does not report that diagnostic.

Strict parsing accepts only the required JSON value with surrounding
whitespace. Markdown fences or additional text remain malformed, even
when the enclosed answer would otherwise be usable.
Valid S/N scores require three answers in $\{1,2,3\}$.
D requires seven probabilities in $[0,1]$ with
$|\sum_s p(s)-1|\leq10^{-3}$. Accepted vectors are divided
by their sum; invalid vectors are not clipped or repaired.
Valid-response and failure rates use all scheduled draws in the
relevant cell or aggregation as the denominator, including failed
draws. The full-study denominator is 9,000.
We distinguish complete valid outputs, item nonresponses,
refusals, malformed outputs, invalid-value outputs,
and technical failures.
 
\subsection{Bootstrap Intervals}
 
We use $B=10{,}000$ iterations. Human respondents are sampled
uniformly with replacement within sample--country--age groups,
retaining group sizes and each score--weight pair.
EU4 and EU27 are resampled separately.
Valid S/N responses and whole D vectors are resampled within
model--country--condition--age--framing cells at their original
valid counts; N has no age index. Framings remain fixed.
 
Each iteration recomputes framing means, equal-framing averages,
gaps, and signed errors. Methods share the same human
resample; both N--age comparisons share the same N resample.
The 2.5th and 97.5th percentiles give pointwise 95\% intervals
\citep{efron1993introduction}. Model resampling conditions on the
selected configurations and valid outputs; human resampling keeps
supplied weights fixed. 
 
Intervals are pointwise rather than simultaneous across comparisons.
An interval containing zero does not establish equivalence or
sufficiently small error.
 
\subsection{Age-Unspecified Baseline}
\label{app:n-baseline}
 
N supplies the same country mean for both age groups.
Its mean absolute error therefore satisfies
\begin{equation}
L_N \geq \frac{|\widehat{\Delta}_H|}{2},
\end{equation}
with equality when its prediction lies between the two human means.
Thus, even the best common prediction cannot reproduce
two distinct age-group means.
 
\section{Additional Model Results}
\label{app:model-results}
 
\subsection{Output Validity}
\label{app:model-validity}
 
All 9,000 scheduled draws were finalized. Table~\ref{tab:model-validity}
records the final classifications. All 72 malformed or
invalid-value outputs were manually reviewed for refusal; none was
identified, and no classification changed. No response was repaired or
regenerated. The four low-validity cells are DeepSeek S-older cells
under framing f2, with 43 (France), 41 (Italy), 44 (Poland), and
42 (Sweden) valid outputs.
 
\begin{table}[!htbp]
\centering\small
\begin{tabular*}{\columnwidth}{@{\extracolsep{\fill}}lr@{}}
\toprule
Final outcome & Draws \\
\midrule
Valid & 8,911 \\
Technical failure & 12 \\
Malformed & 70 \\
Item nonresponse & 5 \\
Invalid values & 2 \\
Refusal & 0 \\
\midrule
Total scheduled & 9,000 \\
\bottomrule
\end{tabular*}
\caption{Final main-run outcomes. All rates use scheduled draws as the denominator.}
\label{tab:model-validity}
\end{table}
 
\subsection{Framing-Specific Simulation Gaps}
\label{app:framing-results}
 
Table~\ref{tab:framing-gaps} shows S gaps before equal-framing aggregation.

\begin{table}[!htbp]
\centering\small
\setlength{\tabcolsep}{3pt}
\begin{tabular}{@{}llrrr@{}}
\toprule
Country & Model & f1 gap & f2 gap & f3 gap \\
\midrule
France & Mistral & 1.245 & 0.260 & 1.280 \\
 & DeepSeek & $-0.451$ & $-1.191^{\dagger}$ & $-1.161$ \\
 & Terra & 0.500 & $-0.820$ & $-0.380$ \\
\addlinespace
Italy & Mistral & 1.347 & 0.820 & 1.525 \\
 & DeepSeek & $-0.611$ & $-1.177^{\dagger}$ & $-0.794$ \\
 & Terra & 0.960 & 0.900 & 0.080 \\
\addlinespace
Poland & Mistral & 1.420 & 1.067 & 1.617 \\
 & DeepSeek & $-0.280$ & $-1.019^{\dagger}$ & $-0.579$ \\
 & Terra & 0.160 & 0.460 & 0.540 \\
\addlinespace
Sweden & Mistral & $-0.376$ & $-0.040$ & 0.510 \\
 & DeepSeek & $-0.560$ & $-0.910^{\dagger}$ & $-0.753$ \\
 & Terra & $-1.700$ & 0.900 & $-0.620$ \\
\bottomrule
\end{tabular}
\caption{Framing-specific S older-minus-younger gaps, rounded to three decimals.
$^{\dagger}$The older-adult cell is a low-validity cell. Point estimates only.}
\label{tab:framing-gaps}
\end{table}
 
\subsection{Human-Reference Sensitivity}
\label{app:reference-results}
 
Table~\ref{tab:reference-sensitivity} compares the same S and D outputs
against EU4 and matched EU27.
 
\begin{table}[!htbp]
\centering\small
\setlength{\tabcolsep}{3pt}
\begin{tabular}{@{}llrrrr@{}}
\toprule
 & & \multicolumn{2}{c}{EU4}
 & \multicolumn{2}{c}{EU27} \\
\cmidrule(lr){3-4}\cmidrule(l){5-6}
Country & Model & S & D & S & D \\
\midrule
France & Mistral & 1.924 & 0.919 & 1.447 & 0.442 \\
 & DeepSeek & $0.062^{\dagger}$ & 0.623 & $0.416^{\dagger}$ & 0.146 \\
 & Terra & 0.762 & 0.043 & 0.285 & 0.434 \\
\addlinespace
Italy & Mistral & 2.124 & 0.939 & 2.583 & 1.398 \\
 & DeepSeek & $0.032^{\dagger}$ & 0.579 & $0.492^{\dagger}$ & 1.038 \\
 & Terra & 1.540 & 0.044 & 1.999 & 0.504 \\
\addlinespace
Poland & Mistral & 2.472 & 0.962 & 2.211 & 0.701 \\
 & DeepSeek & $0.479^{\dagger}$ & 0.698 & $0.217^{\dagger}$ & 0.437 \\
 & Terra & 1.491 & 0.515 & 1.230 & 0.254 \\
\addlinespace
Sweden & Mistral & 1.334 & 1.247 & 1.818 & 1.732 \\
 & DeepSeek & $0.561^{\dagger}$ & 1.071 & $1.046^{\dagger}$ & 1.555 \\
 & Terra & 0.829 & 0.614 & 1.314 & 1.099 \\
\bottomrule
\end{tabular}
\caption{Absolute gap errors under both human references, using unchanged model
outputs. $^{\dagger}$Depends on a low-validity cell. The S/D ranking changes in France for DeepSeek and Terra.}
\label{tab:reference-sensitivity}
\end{table}
 
\subsection{Answer-Order Diagnostic}
\label{app:order-results}
 
A separate S-only screen tested the order in which response options
are displayed. It used France, framing f1, all three models, both age
groups, standard and reversed display orders, and six draws per cell
(72 draws). Reversal changes the display order of codes 1--3 while
preserving code--label mappings and the position of 999; the main run
used the standard order. All 72 screen outputs were valid. The standard
and reversed-order S gaps were, respectively, $1.000$ and $1.333$ for
Mistral, $-0.167$ and $-0.500$ for DeepSeek, and $0.500$ and $-1.667$
for Terra. These point estimates are diagnostic, not precise estimates
of a general order effect.
\end{document}